\documentclass[]{style}

\definecolor{lightblue}{rgb}{0.0, 0.6, 1.0}
\definecolor{darkgreen}{rgb}{0.0, 0.6, 0.0}
\definecolor{lightgreen}{rgb}{0.1, 0.8, 0.1}
\definecolor{lightred}{rgb}{0.7, 0.7, 0.7}
\definecolor{lightgreen}{rgb}{0, 0, 0}
\definecolor{lightlightgray}{rgb}{0.8, 0.8, 0.8}

\usepackage{titlesec}
\titlespacing*{\paragraph}
{0pt}{0em}{0.2em}  

\makeatletter
\renewcommand\paragraph{\@startsection{paragraph}{4}{\z@}%
  {3pt}
  {-0.5em}
  {\normalfont\normalsize\bfseries}} 
\makeatother

\usepackage{amsmath}  
\usepackage{upgreek}
\usepackage{array}    
\usepackage{multirow}
\usepackage{array}
\usepackage{algorithm}
\usepackage{makecell}
\usepackage{graphicx}
\usepackage{algorithm}
\usepackage{algpseudocode}
\usepackage{colortbl}
\usepackage{enumitem}
\usepackage{wrapfig}
\usepackage{mathrsfs}
\usepackage{pifont}
\usepackage{soul}
\usepackage{amssymb}
\usepackage{tcolorbox}
\usepackage{dirtytalk}
\usepackage{xspace}
\usepackage[utf8]{inputenc}

\usepackage{booktabs}

\usepackage{url}
\definecolor{blue}{rgb}{0.21,0.49,0.74}
\definecolor{red}{rgb}{0.8, 0.2, 0.2}
\definecolor{green}{rgb}{0, 0.5, 0}
\definecolor{yellow}{RGB}{218, 160, 109}
\definecolor{gray}{RGB}{155, 155, 155}

\crefname{section}{Sec.}{Secs.}
\Crefname{section}{Section}{Sections}
\Crefname{table}{Table}{Tables}
\crefname{table}{Tab.}{Tabs.}
\crefname{figure}{Fig.}{Figs.}
\Crefname{figure}{Figure}{Figures}
\crefname{appendix}{App.}{Apps.}
\Crefname{appendix}{Appendix}{Appendices}

\usepackage[utf8]{inputenc} 
\usepackage[T1]{fontenc}    
\usepackage{graphicx}
\usepackage{amsmath}
\usepackage{amssymb}
\usepackage{url}            
\usepackage{booktabs}       
\usepackage{amsfonts}       
\usepackage{nicefrac}       
\usepackage{microtype}      
\usepackage{xcolor}         
\usepackage{multirow}
\usepackage{comment}
\usepackage{colortbl}
\usepackage{tocloft}
\usepackage{algorithm}
\usepackage{algorithmicx}
\usepackage{algpseudocode}
\usepackage{wrapfig}
\usepackage{tabularx}
\usepackage{textcomp}
\usepackage{stfloats}
\usepackage{verbatim}
\usepackage{titlesec}
\usepackage{adjustbox}
\usepackage{pifont}
\usepackage{tikz}
\usepackage{color}
\usepackage{subcaption}
\usepackage{soul}
\usepackage{mathtools}

\usepackage{makecell}
\newcolumntype{Y}{>{\centering\arraybackslash}X}
\usepackage{multirow}
\usepackage[table]{xcolor} 
\definecolor{bestcolor}{HTML}{A9D18E} 
\definecolor{sbestcolor}{HTML}{E2EFDA}

\newcolumntype{C}{>{\centering\arraybackslash}m{1.3cm}}
\newcolumntype{Z}{>{\raggedright\arraybackslash\fontsize{6.4pt}{7.2pt}\selectfont}X}

\RequirePackage{xspace}
\makeatletter
\DeclareRobustCommand\onedot{\futurelet\@let@token\@onedot}
\def\@onedot{\ifx\@let@token.\else.\null\fi\xspace}

\makeatother

\definecolor{lightblue}{rgb}{0.66, 0.85, 0.95}
\hypersetup{
    colorlinks=true,
    linkcolor=red,
    citecolor=blue,
}
\definecolor{c2}{HTML}{FBD9BD}
\definecolor{c3}{HTML}{fe793d}
\definecolor{c4}{HTML}{eedeb0}
\definecolor{rouse}{rgb}{0.981,0.961,0.941}
\usepackage[most]{tcolorbox}
\tcbset{colback=blue!5!white, colframe=blue!20!white, boxrule=0pt, arc=2mm, left=1mm, right=1mm, top=1mm, bottom=1mm}

\definecolor{adptorange}{RGB}{248, 205, 172}
\definecolor{cmpblue}{RGB}{189, 215, 238}
\definecolor{cmpblue}{RGB}{189, 215, 238}

\definecolor{our_red}{RGB}{232,157,160}
\definecolor{our_blue}{RGB}{136,206,230}
\definecolor{our_orange}{RGB}{246,200,168}
\definecolor{our_green}{RGB}{178,211,164}

\definecolor{attn_code0}{RGB}{247,215,200}
\definecolor{attn_code1}{RGB}{238,169,139}
\definecolor{mlp_code0}{RGB}{204,201,221}
\definecolor{mlp_code1}{RGB}{102,95,153}

\definecolor{token_blue}{RGB}{84, 120, 140}

\usepackage{pifont}       
\usepackage{bbding}       
\usepackage{fontawesome}
\usepackage{xspace}

\usepackage{float}
\usepackage{placeins}

\newlength\savewidth

\newcommand\nnfootnote[1]{
  \begin{NoHyper}
  \renewcommand\thefootnote{}\footnote{#1}
  \addtocounter{footnote}{-1}
  \end{NoHyper}
}

\newcolumntype{x}[1]{>{\centering\arraybackslash}p{#1pt}}
\newcolumntype{y}[1]{>{\raggedright\arraybackslash}p{#1pt}}
\newcolumntype{z}[1]{>{\raggedleft\arraybackslash}p{#1pt}}

\renewcommand{\paragraph}[1]{\vspace{1mm}\noindent\textbf{#1}}

\usepackage{colortbl}
\usepackage{xcolor}
\usepackage{wrapfig}

\renewcommand{\paragraph}[1]{\vspace{1.25mm}\noindent\textbf{#1}}

\usepackage{listings}

\definecolor{codeblue}{rgb}{0.21, 0.49, 0.74}
\definecolor{codekw}{rgb}{0.35, 0.35, 0.75}
\lstdefinestyle{Pytorch}{
    language = Python,
    backgroundcolor = \color{white},
    basicstyle = \fontsize{9pt}{8pt}\selectfont\ttfamily\bfseries,
    columns = fullflexible,
    aboveskip=1pt,
    belowskip=1pt,
    breaklines = true,
    captionpos = b,
    commentstyle = \color{codeblue},
    keywordstyle = \color{codekw},
}

\definecolor{green}{HTML}{009000}
\definecolor{red}{HTML}{ea4335}

\title{MoVerse: Real-Time Video World Modeling with Panoramic Gaussian Scaffold}
\author[]{Yang Zhou\textsuperscript{$13{\ast}{\dagger}$}}
\author[]{Ziheng Wang\textsuperscript{$23{\ast}{\dagger}$}}
\author[]{Yuqin Lu\textsuperscript{$13{\ast}{\dagger}$}}
\author[]{Haofeng Liu\textsuperscript{$3{\ddagger}$}}
\author[]{Jun Liang\textsuperscript{$3{\ddagger}$}}
\author[]{Shengfeng He\textsuperscript{$4$}}
\author[]{Jing Li\textsuperscript{$3$}}

\affiliation[1]{South China University of Technology}
\affiliation[2]{Columbia University}
\affiliation[3]{Orange Team, Moku Lab, HUJING Digital Media \& Entertainment Group}
\affiliation[4]{Singapore Management University}

\abstract{
We present MoVerse, a real-time video world model that creates an interactively navigable scene from a single narrow-field-of-view image. This setting is challenging because the input observes only a small fraction of the environment, while interactive roaming requires a complete surrounding world, persistent geometry, controllable camera motion, and temporally coherent high-fidelity observations. MoVerse addresses this problem by separating world construction from observation rendering. It first expands the input into a gravity-aligned 360$^\circ$ panorama with topology-aware diffusion, closing the missing field of view before 3D reasoning. It then lifts the panorama into a persistent 3D Gaussian scaffold using panoramic geometry-aware residual prediction, yielding a dense and directly renderable spatial memory. Finally, a Gaussian-conditioned video renderer translates scaffold renderings along user-specified camera trajectories into photorealistic video. To make this renderer practical for interaction, we train a bidirectional diffusion teacher for high-quality conditional rendering and distill it into a causal autoregressive student for bounded-latency streaming. This design combines the controllability and long-range consistency of explicit 3D representations with the perceptual quality of generative video models. MoVerse supports real-time scene roaming at 8~FPS on a single NVIDIA RTX~4090 GPU, demonstrating a practical path toward single-image world creation with interactive video output.
}

\project{\url{https://orange-3dv-team.github.io/MoVerse/}}
\date{\today}

\begin{document}
\thispagestyle{firstheader}
\maketitle

\nnfootnote{$\ast$ Work done during an internship at Orange Team, Moku Lab, HUJING Digital Media \& Entertainment Group.}
\nnfootnote{$\dagger$ Equal contribution.}
\nnfootnote{$\ddagger$ Project leader.}

\begin{center}
  \includegraphics[width=\linewidth]{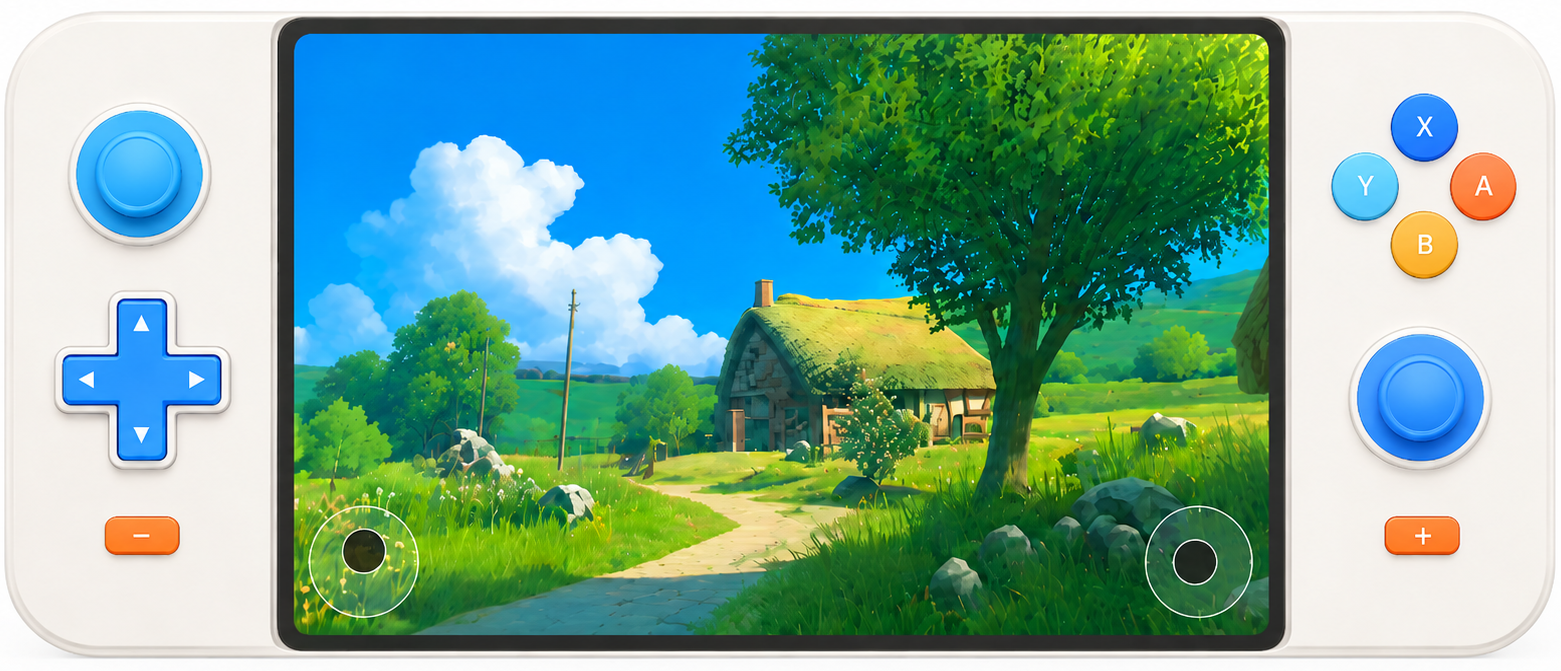}
\end{center}

\pagestyle{plain}

\section{Introduction}
\label{sec:intro}

Generating a navigable world from a single narrow-field-of-view (NFOV) image is fundamentally under-constrained. The input observes only a small frustum, while interactive applications such as VR prototyping, digital twins, content creation, and embodied-agent simulation require a user to move through a complete, spatially persistent environment. A practical system must therefore solve three coupled problems: it must complete the missing field of view, convert that completion into a camera-controllable scene representation, and render high-quality temporally coherent video during interaction.

Existing approaches usually emphasize only part of this requirement. Explicit 3D methods build persistent scene assets, such as point clouds, meshes, or 3D Gaussian scenes, and then render novel views from them~\cite{yu2024wonderjourney, yu2025wonderworld, yang2025layerpano3d, zheng2025self, team2025hunyuanworld, schneider2025worldexplorer, yang2025matrix, shen2026lyra, hy2026hy}. Their explicit state provides durable spatial memory, accurate camera control, and high frame-rate rendering. However, when the asset is lifted directly from a single NFOV image, most of the scene must be inferred from weak evidence; when additional views are synthesized before reconstruction, the resulting asset can inherit cross-view inconsistencies and generation cost. Direct rendering from such assets may expose holes, floaters, depth errors, or limited perceptual quality, especially under large viewpoint changes.

Implicit video and world models take the opposite route~\cite{genie3, rtfm2025, yu2025context, he2025matrix, hyworld2025, mao2025yume, hong2025relic, team2026advancing, wang2026matrix, zhu2026sana}. They generate observations during interaction and store history in attention windows, recurrent states, or key--value caches. These models can produce visually strong videos, but their long-range geometry is only as stable as the implicit memory retained by the model. As the user follows a long trajectory or revisits a previously seen region, the scene can drift, change identity, or develop boundary discontinuities that are difficult to correct online.

Hybrid systems use explicit geometry as a condition for generative rendering~\cite{wang2025evoworld, ren2025gen3c, yu2025trajectorycrafter, liu2026mocam, wang2026one2scene, team2026inspatio}. They show that a geometric anchor can constrain camera motion and scene layout while a learned renderer improves appearance. However, the strength of the anchor and the deployability of the renderer remain in tension: sparse or coarse anchors give limited spatial guidance, whereas strong generative renderers are often too expensive or too bidirectional for real-time interaction. This motivates a design in which the explicit condition is dense, panoramic, directly renderable, and reusable, while the video model is distilled into a causal renderer for streaming use.

We present \textbf{MoVerse}, a three-stage system for real-time video world modeling with panoramic Gaussian guidance. Given a single NFOV image, MoVerse first generates a gravity-aligned 360$^\circ$ equirectangular panorama, then lifts the panorama into a persistent 3D Gaussian scaffold, and finally renders interactive video through a Gaussian-conditioned autoregressive video model. The key design choice is to separate world construction from observation rendering: Stages~I and~II build a reusable scaffold offline, while Stage~III turns scaffold renderings along user-specified camera trajectories into high-fidelity video online.

MoVerse is organized around three technical components:
\begin{itemize}[leftmargin=*, itemsep=2pt, topsep=2pt]
  \item \textbf{Geometry-aware panoramic generation.} Stage~I closes the missing field of view before 3D lifting. It canonicalizes an unposed NFOV input into a gravity-aligned panoramic frame using differentiable auto-leveling, then performs masked latent diffusion completion with circular latent encoding and shift-equivariant generation to respect the horizontal $S^1$ topology of ERP panoramas. The model is trained on Horizon360, a curated set of canonicalized panoramas with yaw-centered perspective-view supervision.
  \item \textbf{Panoramic Gaussian scaffold construction.} Stage~II converts the completed panorama into a directly renderable 3DGS scene using a feed-forward panoramic Gaussian generator. The scaffold is initialized by spherical back-projection from panorama depth, uses latitude-aware scale correction proportional to $\cos\phi$, and predicts residuals in angular--inverse-depth space so that Gaussian updates remain consistent with ERP geometry and horizontal closure.
  \item \textbf{Gaussian-conditioned streaming video rendering.} Stage~III renders the scaffold along the requested camera trajectory and translates the resulting RGB conditioning stream into final video. A bidirectional teacher, initialized from Wan2.1-T2V, learns dense Gaussian-conditioned video rendering with shared positional coordinates between target and condition tokens. It is then distilled into a causal autoregressive student using self-forcing and distribution matching, enabling bounded-latency streaming with a local key--value cache while the explicit scaffold carries long-range spatial memory.
\end{itemize}

This factorization gives each stage a clear role. The panorama generator supplies complete omnidirectional evidence instead of asking the 3D module to hallucinate most of the world. The Gaussian predictor turns that evidence into a persistent renderable asset rather than leaving scene state inside a video model. The causal renderer improves perceptual quality and temporal coherence without replacing the camera motion and layout encoded by the scaffold. In our deployment configuration, this design supports real-time roaming at 8~FPS on a single NVIDIA RTX~4090 GPU.

\paragraph{Stage-wise positioning.}
The three stages connect several lines of prior work. Stage~I builds on panorama outpainting and 360$^\circ$ generation~\cite{wupanodiffusion, feng2025dit360, zheng2025panorama, yuan2025camfreediff}, but emphasizes gravity canonicalization and spherical topology. Stage~II follows feed-forward 3DGS reconstruction~\cite{jiang2025anysplat, kim2025vg3t, chen2025splatter, zhang2025pansplat, ren2025panosplatt3r, mescheder2025sharp}, but specializes residual Gaussian prediction to ERP panoramas. Stage~III adopts the conditioning pattern of camera-controlled video and novel-view synthesis~\cite{liu2026mocam, ren2025gen3c, yu2025trajectorycrafter} on top of modern video generators~\cite{wan2025wan, yang2024cogvideox, kong2024hunyuanvideo}, realizing causal streaming rather than offline clip refinement.
\section{Method}

\subsection{Overview}
\begin{figure}[t]
  \centering
  \includegraphics[width=\linewidth]{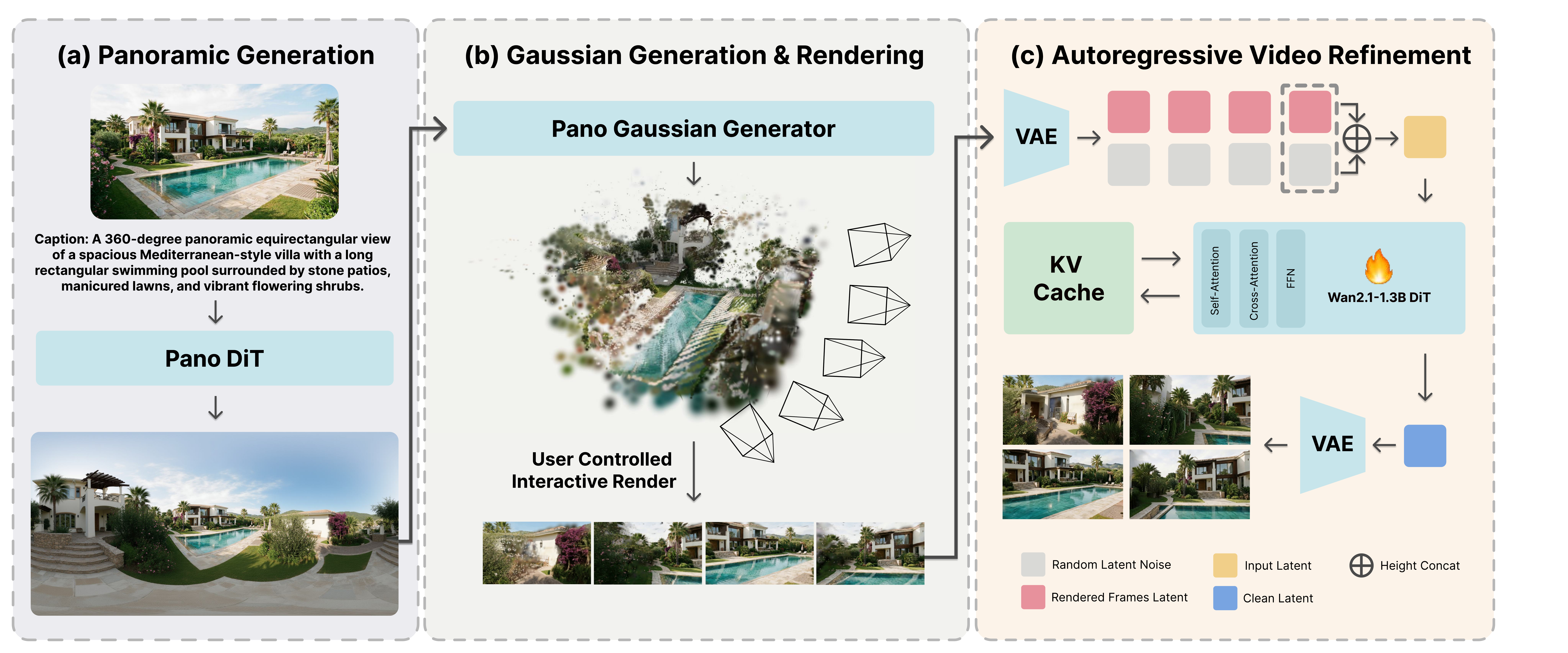}
  \caption{\textbf{MoVerse pipeline overview.} From a single narrow-field-of-view input image, Stage~I synthesizes a gravity-aligned 360$^\circ$ panorama, Stage~II lifts the panorama into a persistent 3D Gaussian scaffold, and Stage~III renders Gaussian-conditioned video along user-specified camera trajectories for real-time interactive roaming.}
  \label{fig:pipeline_overview}
\end{figure}

Given a single NFOV image $I \in \mathbb{R}^{H \times W \times 3}$, MoVerse produces a real-time navigable video stream through three sequential stages, as illustrated in Fig.~\ref{fig:pipeline_overview}. The pipeline first completes the missing omnidirectional context, then stores the completed world in a persistent Gaussian scaffold, and finally renders interactive observations by translating scaffold renderings into temporally coherent video.

\begin{enumerate}[leftmargin=*, itemsep=2pt, topsep=2pt]
  \item \textbf{Stage~I: Panoramic Generation} ($I \to P$). The input image is expanded into a gravity-aligned 360$^\circ$ equirectangular panorama $P$. This step closes the missing field of view \emph{before} any 3D lifting, providing structurally complete observations for every viewing direction and eliminating the need for the subsequent 3D stage to hallucinate large unobserved regions.

  \item \textbf{Stage~II: Gaussian Scene Generation} ($P \to \mathcal{G}$). The panorama is lifted into a panoramic 3D Gaussian scene $\mathcal{G} = \{(\mu_k, \Sigma_k, \alpha_k, c_k)\}_{k=1}^{K}$, where each Gaussian is parameterized by its center $\mu_k$, covariance $\Sigma_k$, opacity $\alpha_k$, and appearance $c_k$. Unlike point-cloud or mesh representations, 3D Gaussians are directly splatting-renderable at real-time frame rates, yielding a dense, camera-controllable RGB conditioning video $\hat{V}_{1:T}=\{\hat{V}_t\}_{t=1}^{T}$ that is substantially more informative than sparse point-cloud projections, but still imperfect as final photorealistic observations.

  \item \textbf{Stage~III: Gaussian-Conditioned Video Rendering} ($\hat{V}_{1:T} \to V_{1:T}$). A learned video renderer translates the Gaussian-rendered conditioning video $\hat{V}_{1:T}$ into a high-fidelity output video $V_{1:T}=\{V_t\}_{t=1}^{T}$. By generating \emph{sequences} rather than independent frames, the renderer maintains temporal coherence across successive viewpoints while enhancing perceptual quality without changing the camera motion or scene layout implied by the scaffold.
\end{enumerate}

\paragraph{Offline vs.\ online split.}
Stages~I and~II are executed \emph{once} as offline scaffold construction: given a new input image, they produce a persistent 3DGS asset in seconds. Stage~III operates \emph{online}: at interaction time, the scaffold is rendered at the user-requested camera trajectory, and the video renderer streams enhanced frames in real time. This separation ensures that the computationally heavier generative steps (panorama synthesis, 3D lifting) do not bottleneck interactive exploration, while the explicit scaffold provides durable geometric memory that prevents drift across arbitrarily long trajectories.

\subsection{Stage I: Panoramic Generation}\label{sec:stage1}

\begin{figure}[t]
  \centering
  \includegraphics[width=\linewidth]{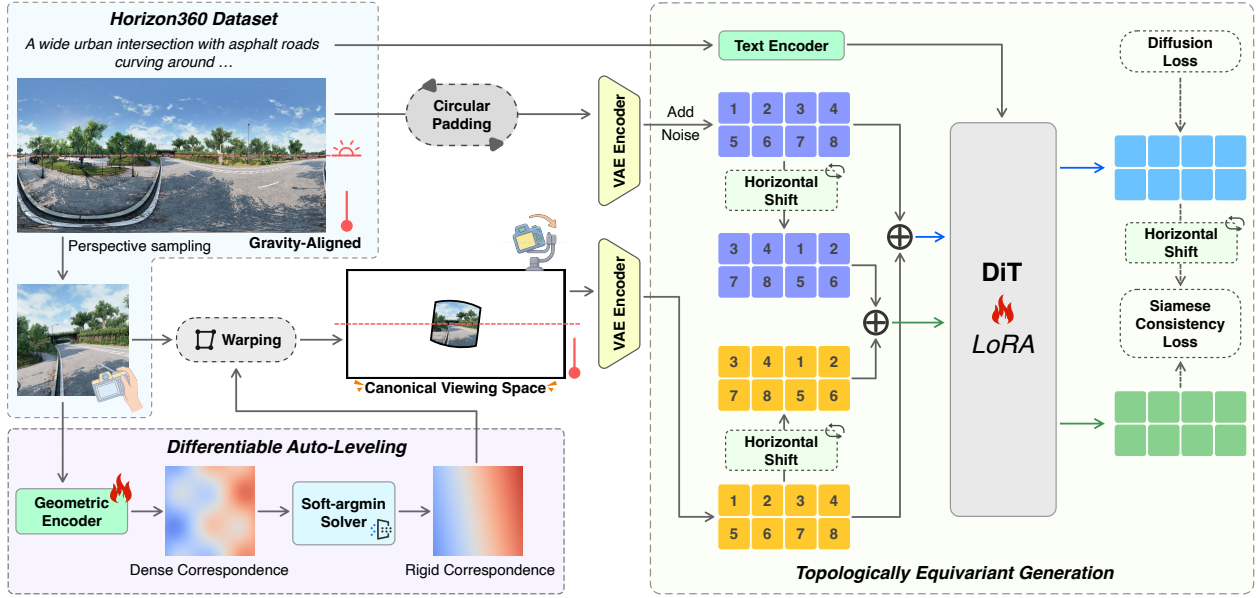}
  \caption{\textbf{Stage~I panoramic generation.} The input NFOV image is auto-leveled into a gravity-aligned canonical viewing space and then completed as a topology-aware ERP panorama. Circular latent encoding and shift-equivariant generation preserve the horizontal $S^1$ boundary.}
  \label{fig:stage1_framework}
\end{figure}

Stage~I expands the input NFOV image $I$ into a complete equirectangular panorama $P \in \mathbb{R}^{H_e \times W_e \times 3}$. Image outpainting and completion have progressed from GAN-based extrapolation to diffusion-based inpainting and latent completion~\cite{zheng2019pluralistic,wang2019wide,zhao2021large,rombach2022high,lugmayr2022repaint}. In parallel, recent panoramic generation methods have adapted diffusion priors to $360^\circ$ content synthesis, including text-to-panorama generation, perspective-conditioned panorama outpainting, and immersive world generation~\cite{li2023panogen,feng2023diffusion360,wupanodiffusion,zheng2025panorama,feng2025dit360,team2025hunyuanworld,lu2025matrix3d}. Rather than treating this step as generic image outpainting, we formulate it as geometry-aware panoramic completion. The generated panorama should provide complete $360^\circ$ scene context, maintain a gravity-aligned horizon, and respect the horizontal $S^1$ periodicity of ERP images. We denote this stage as
\begin{equation}
P = G_{\theta}(I),
\end{equation}
where $G_{\theta}$ first canonicalizes the input view and then completes the surrounding omnidirectional scene. Fig.~\ref{fig:stage1_framework} summarizes the resulting Stage~I pipeline: the input is stabilized into a canonical panoramic frame, used as masked visual context for diffusion-based completion, and decoded with topology-aware operations to produce a horizontally closed ERP panorama.

The generator is implemented as a conditional latent diffusion completion model, following the common practice of performing high-resolution synthesis in a compressed latent space~\cite{rombach2022high}. Let $\mathcal{E}$ and $\mathcal{D}$ denote the latent encoder and decoder. During training, a canonical target panorama is encoded as $z_0 = \mathcal{E}(P)$, and a binary ERP mask $M$ marks the known region induced by the sampled input view. The visible reference latent is
\begin{equation}
z_{\mathrm{ref}} = z_0 \odot M.
\end{equation}
where $\odot$ denotes element-wise multiplication, with $M$ broadcast across latent channels. At inference time, the known latent context is instead formed from the input image after projection and canonical alignment. During denoising, the network predicts the injected noise from the noisy latent, the mask, and the known visual context:
\begin{equation}
\hat{\epsilon}
= \epsilon_{\theta}(z_t \oplus M \oplus z_{\mathrm{ref}}, t, \tau(y)),
\end{equation}
where $\oplus$ denotes channel-wise concatenation and $\tau(y)$ is the conditioning embedding. The corresponding latent diffusion objective is
\begin{equation}
\mathcal{L}_{\mathrm{LDM}}
= \mathbb{E}_{z_0,\epsilon,t}
\left[
\left\|\epsilon - \epsilon_{\theta}(z_t \oplus M \oplus z_{\mathrm{ref}}, t, \tau(y))\right\|_2^2
\right].
\end{equation}
This masked latent formulation preserves the observed input while synthesizing the unobserved panoramic field of view.

A direct perspective-to-ERP outpainting pipeline is insufficient for this purpose because it ignores two domain gaps. The first is \emph{projective variance}. Real input images are generally unposed: arbitrary pitch and roll distort the projected ERP observation, causing straight walls, floor boundaries, and horizons to become curved or tilted. Many panorama outpainting pipelines project the input view into ERP using predefined camera assumptions or known extrinsics~\cite{wupanodiffusion,zheng2025panorama,feng2025dit360,team2025hunyuanworld,lu2025matrix3d}, which limits robustness for unconstrained photographs. These non-linear distortions conflict with the structural priors that image diffusion models learn from perspective imagery. The second is \emph{topological severing}. ERP panoramas are horizontally periodic: the left and right image boundaries correspond to adjacent azimuthal directions on the same physical sphere. Standard diffusion backbones, VAEs, and positional encodings are designed for bounded Euclidean images and can therefore introduce artificial seams at this boundary. Existing works mitigate this issue through circular blending, synchronized denoising, circular padding, or specialized panoramic operators~\cite{feng2023diffusion360,lee2023syncdiffusion,wang2024360dvd,liao2023cylin}, but these treatments do not fully remove the underlying mismatch between Euclidean image grids and the closed azimuthal topology. We address these two gaps in order: first by defining a gravity-aligned canonical panoramic representation, and then by enforcing topology-aware completion in the latent generation process.

\subsubsection{Canonical panoramic representation.}
We complete the scene in a canonical viewing space $\mathcal{C}$. This space removes pitch and roll ambiguity by aligning the environmental horizon with the equator of the ERP canvas. The output of Stage~I is therefore not merely a visually plausible panorama, but a normalized panoramic observation with stable longitude--latitude coordinates. In this representation, vertical architectural structures remain plumb and the horizon remains level. A separate yaw-centered target convention, described below in the training data construction, places the input view at the center of the panorama rather than near the periodic boundary. We write the gravity canonicalization abstractly as
\begin{equation}
z_{\mathrm{can}} = \mathcal{T}_{\mathrm{align}}(z_{\mathrm{ref}}),
\end{equation}
where $\mathcal{T}_{\mathrm{align}}$ maps the observed latent context into the gravity-aligned panoramic frame.

\subsubsection{Differentiable auto-leveling.}
We adopt differentiable auto-leveling to estimate $\mathcal{T}_{\mathrm{align}}$ for uncalibrated perspective inputs without requiring camera extrinsics at inference time. The auto-leveling module first predicts a dense 2D correspondence field
\begin{equation}
P_{\mathrm{dense}} \in \mathbb{R}^{H_I \times W_I \times 2},
\end{equation}
which represents pixel-wise shifts toward a gravity-leveled reference. Instead of directly using this dense field as a free-form spatial warp, the correspondences are passed through a rigid transformation bottleneck implemented by a differentiable soft-argmin solver. The solver searches for a low-dimensional rigid spherical transformation $\hat{R}$ whose reprojection best explains the dense correspondences, including the gravity-alignment rotation needed to remove pitch and roll under the canonical yaw convention. The actual canonical warp is then performed by spherical grid sampling:
\begin{equation}
 z_{\mathrm{can}} = \mathcal{W}(z_{\mathrm{ref}}, \hat{R}),
\end{equation}
where $\mathcal{W}$ denotes differentiable spherical sampling.

This constraint is important for stable diffusion-based training. Although spatial transformer networks provide a general mechanism for differentiable image warping~\cite{jaderberg2015spatial}, an unconstrained spatial transformer can absorb high-frequency reconstruction gradients as local non-rigid distortions, producing unstable jelly-like warps. The rigid bottleneck instead restricts alignment updates to globally coherent camera rotations. To stabilize early training, we also apply an auxiliary geometric anchor to the dense correspondence prediction:
\begin{equation}
\mathcal{L}_{\mathrm{flow}}
= \mathrm{Smooth}_{L_1}(P_{\mathrm{dense}}, P_{\mathrm{GT}}),
\end{equation}
where $P_{\mathrm{GT}}$ is analytically obtained from canonicalized panorama--view pairs. The final warp still passes through the rigid solver, so the dense field supervises alignment evidence rather than becoming a free-form deformation field.

\subsubsection{Topology-aware panoramic completion.}
After canonicalization, the generator completes the missing field of view while preserving the horizontal closure of the ERP domain. We use circular latent encoding so that horizontal convolutional padding in the latent autoencoder wraps around the azimuthal axis rather than introducing artificial image borders. This makes the latent compression and reconstruction process consistent with the periodic condition of ERP panoramas.

We further encourage shift-equivariant diffusion generation. Let $\mathrm{Roll}_{\delta}(\cdot)$ denote a horizontal circular shift by offset $\delta$. For the canonical diffusion input $X = z_t \oplus M \oplus z_{\mathrm{can}}$, the base and shifted branches are
\begin{equation}
\hat{\epsilon}_{\mathrm{base}}
= \epsilon_{\theta}(X, t, \tau(y)),
\end{equation}
\begin{equation}
\hat{\epsilon}_{\mathrm{shifted}}
= \epsilon_{\theta}(\mathrm{Roll}_{\delta}(X), t, \tau(y)).
\end{equation}
The shift-consistency objective is
\begin{equation}
\mathcal{L}_{\mathrm{shift}}
= \mathbb{E}_{z_0, \epsilon, t, \delta}
\left[
\left\|
\mathrm{Roll}_{\delta}(\hat{\epsilon}_{\mathrm{base}})
- \hat{\epsilon}_{\mathrm{shifted}}
\right\|_2^2
\right].
\end{equation}
For transformer-based diffusion backbones, this circular shift can be implemented efficiently by shifting positional coordinates rather than physically rolling all latent tokens. During inference, random horizontal shifts can also be applied across denoising steps and inverted at the end, so no fixed tensor boundary consistently acts as the seam location.

\subsubsection{Canonical training data.}
\begin{figure}[t]
  \centering
  \includegraphics[width=\linewidth]{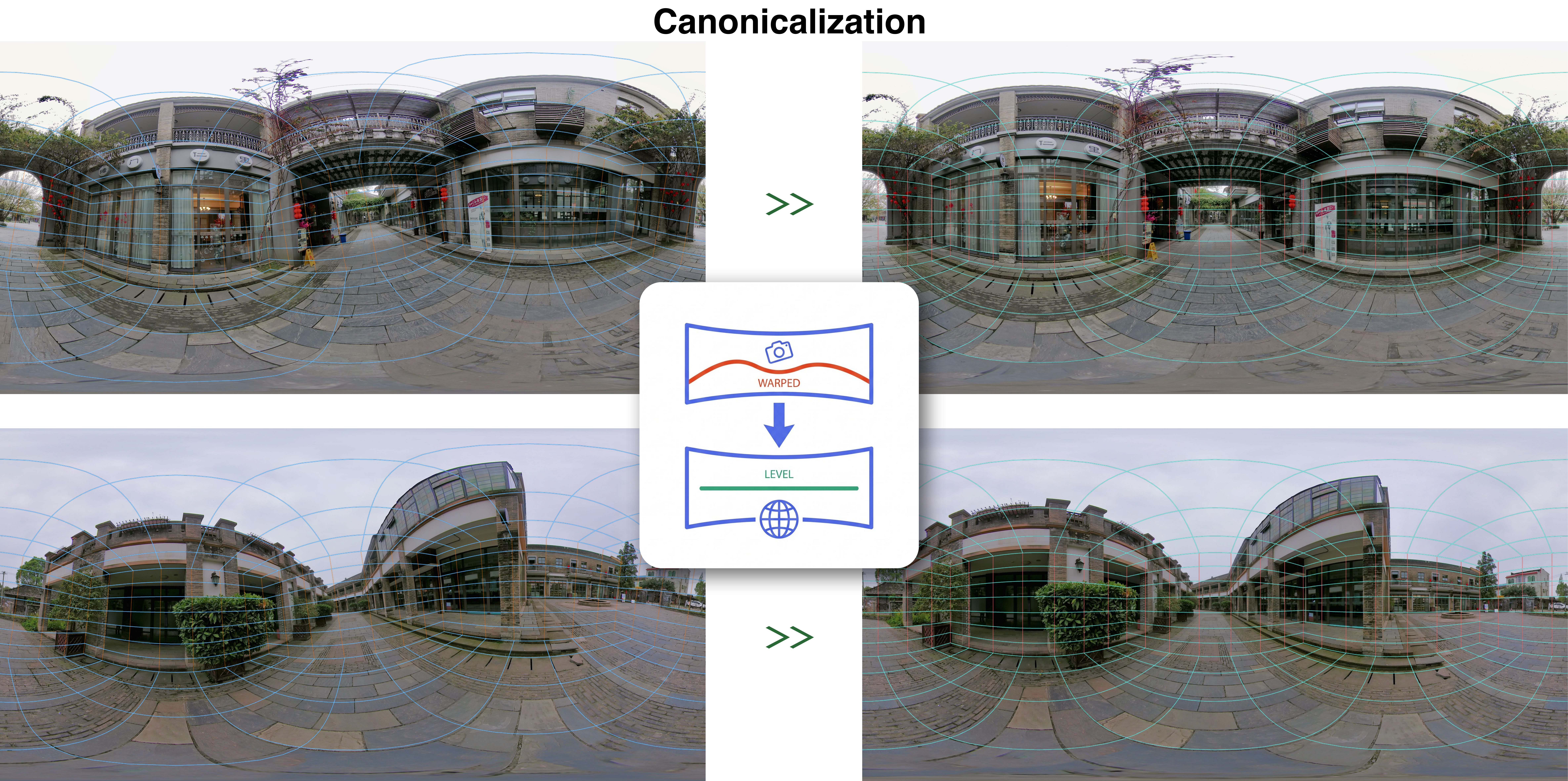}
  \caption{\textbf{Canonical panorama construction.} Raw unaligned panoramas are rotated on the sphere so that the horizon coincides with the ERP equator and physical verticals become plumb, yielding the canonical training targets used by Stage~I.}
  \label{fig:stage1_canonicalization}
\end{figure}

The panoramic generator is trained with Horizon360, a curated collection of gravity-aligned indoor and outdoor panoramas. Existing omnidirectional datasets and scene resources provide diverse panoramic content~\cite{xiao2012recognizing,chang2017matterport3d,polyhaven,yang2025layerpano3d}, but raw panoramas often contain inconsistent pitch, roll, and horizon placement. The raw panoramas are therefore canonicalized by rotating the spherical image so that the true horizon aligns with the ERP equator and physical verticals remain plumb, following the geometric intuition behind panoramic layout and horizon estimation~\cite{zhang2014panocontext,zou2018layoutnet,sun2019horizonnet,jiang2022lgt,sun2021hohonet}. Fig.~\ref{fig:stage1_canonicalization} visualizes this rectification process. This provides the diffusion model with a consistent structural prior instead of asking it to learn from panoramas with arbitrary tilt and roll.

To expose the model to realistic input photographs, perspective views are sampled from the canonical panoramas with variable yaw, pitch, roll, field of view, and aspect ratio. The sampling distribution mixes common handheld captures with more extreme camera poses, so the auto-leveling module learns both small everyday misalignments and challenging out-of-distribution tilts. We also circularly roll each target panorama by the input yaw so that the observed view is centered in the canonical canvas:
\begin{equation}
P_{\mathrm{centered}} = \mathrm{Roll}(P_{\mathrm{ERP}}, -\psi_{\mathrm{input}}).
\end{equation}
This yaw-centered convention keeps the known region away from the periodic boundary and makes completion more stable.

The Stage~I training objective combines the standard latent diffusion loss with the geometric and topological regularizers:
\begin{equation}
\mathcal{L}_{\mathrm{stage1}}
= \mathcal{L}_{\mathrm{LDM}}
+ \lambda_{\mathrm{shift}}\mathcal{L}_{\mathrm{shift}}
+ \lambda_{\mathrm{flow}}\mathcal{L}_{\mathrm{flow}}.
\end{equation}
The final output $P = G_{\theta}(I)$ is an equirectangular, gravity-aligned, horizontally periodic panorama centered around the input view, which serves as the canonical panoramic input for the following 3D construction stage.
\subsection{Stage II: Gaussian Scene Generation}\label{sec:stage2}

\begin{figure}[t]
  \centering
  \includegraphics[width=\linewidth]{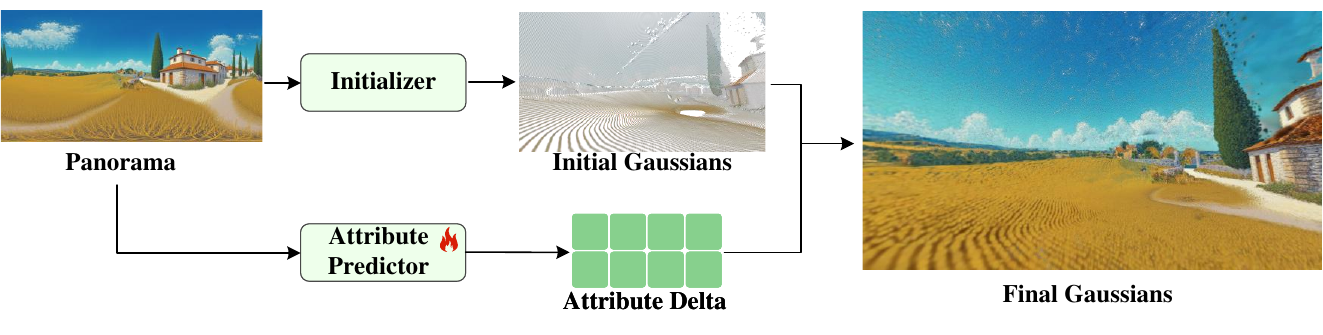}
  \caption{\textbf{Stage~II Gaussian scene generation.} Given the completed gravity-aligned panorama from Stage~I, MoVerse initializes Gaussian primitives in the panoramic domain with latitude-aware scaling, and predicts residual Gaussian attributes in angular--inverse-depth space to obtain a persistent 3DGS scaffold for downstream video rendering.}
  \label{fig:stage2_pipeline}
\end{figure}

Stage~II converts the gravity-aligned panorama $P$ from Stage~I into a persistent, directly renderable 3D Gaussian scaffold $\mathcal{G}$, building a dense and stable geometry-and-appearance anchor for downstream synthesis. We therefore utilize a feed-forward 3DGS predictor: it runs once per input panorama, produces an explicit representation that can be reused during interaction, and yields informative RGB conditioning frames for the video renderer. Fig.~\ref{fig:stage2_pipeline} summarizes this panorama-to-Gaussian pathway: the model constructs an ERP-aware spherical Gaussian initialization and then predicts residual corrections to geometry and appearance before producing a standard splatting-renderable 3DGS asset.

\subsubsection{Panoramic Gaussian initialization.}
The input panorama is represented in equirectangular projection (ERP). Each pixel $(u,v)$ corresponds to a longitude--latitude pair
\begin{equation}
\theta = \left(\frac{u}{W}-\frac{1}{2}\right)2\pi,
\qquad
\phi = \left(\frac{1}{2}-\frac{v}{H}\right)\pi,
\end{equation}
and hence to a unit viewing direction
\begin{equation}
\mathbf{d}(\theta,\phi)=
\bigl(\cos\phi\sin\theta,\; -\sin\phi,\; \cos\phi\cos\theta\bigr).
\end{equation}
Given an estimated panorama depth map $D$, we instantiate Gaussians on a strided ERP grid $\Omega'=\{(u_k,v_k)\}_{k=1}^{K}$ with resolution $H'\times W'$, so $K=H'W'$. For each grid location, let
\begin{equation}
\theta_k=\theta(u_k),
\qquad
\phi_k=\phi(v_k),
\qquad
\mathbf{d}_k=\mathbf{d}(\theta_k,\phi_k),
\qquad
D_k=D(u_k,v_k).
\end{equation}
The initial Gaussian center is obtained by spherical back-projection,
\begin{equation}
\mu_k = D_k\mathbf{d}_k.
\end{equation}
We also initialize the Gaussian scale in an ERP-aware manner. Since the spherical area represented by an ERP pixel is proportional to $\cos\phi_k$, the tangential footprint associated with longitude spacing should shrink toward high latitudes. We therefore set the initial scale as
\begin{equation}
s_k \propto D_k\cos\phi_k,
\end{equation}
with a lower bound to avoid degenerate splats near the poles. Each primitive then carries a center, anisotropic scale, rotation, color, and opacity, yielding
\begin{equation}
\mathcal{G}=\{(\mu_k, s_k, q_k, c_k, \alpha_k)\}_{k=1}^{K},
\end{equation}
where $\mu_k$ is the Gaussian center, $s_k$ denotes the 3D scale vector, $q_k$ a quaternion rotation, $c_k$ RGB appearance, and $\alpha_k$ opacity.

\subsubsection{Depth-guided residual prediction.}
Our panoramic Gaussian generator follows the depth-guided residual 3DGS prediction principle of SHARP~\cite{mescheder2025sharp}, but adapts it from perspective images to equirectangular panoramas. This adaptation replaces planar camera assumptions with spherical geometry: pixel displacements correspond to angular changes, sampling density varies with latitude, and the horizontal boundary must remain topologically closed. Starting from the initialized scaffold, we extract multi-scale image features with a pretrained vision backbone and decode them with a DPT-style feature head into geometry and texture features. The feature head processes the encoder features from the panorama depth model and further fuses an explicit feature input formed by concatenating the RGB panorama with inverse depth, i.e., $[P, D^{-1}]$. Let $\mathbf{f}_k$ denote the decoded feature vector at grid location $k$, and let $h_\psi$ denote the lightweight residual prediction head with learnable parameters $\psi$. The head predicts residual Gaussian parameters as
\begin{equation}
(\Delta\theta_k,\Delta\phi_k,\Delta z_k,\Delta s_k,\Delta q_k,\Delta c_k,\Delta\alpha_k)
= h_\psi(\mathbf{f}_k).
\end{equation}
The prediction head is zero-initialized, so before learning the network reproduces the initialized spherical scaffold. Training therefore starts from a physically meaningful initialization and only learns local corrections. This residual formulation is intended to keep the scaffold anchored to the panorama and the provided depth geometry, while still allowing the network to correct local errors needed for multi-view renderability.

\subsubsection{Attribute composition.}
We compose all predicted Gaussian quantities as residuals around the initialized scaffold. The most important case is the Gaussian center, because ERP panoramas use an angular image domain rather than a planar one. In a pinhole camera, image-plane offsets are naturally expressed in normalized device coordinates. In an ERP panorama, horizontal motion changes longitude and vertical motion changes latitude. For each primitive $k$, we therefore predict position offsets $(\Delta\theta_k,\Delta\phi_k,\Delta z_k)$ and apply them in angular--inverse-depth space before spherical back-projection:
\begin{equation}
\theta'_k = \theta_k + \lambda_{xy}\Delta\theta_k,
\qquad
\phi'_k = \mathrm{clip}(\phi_k + \lambda_{xy}\Delta\phi_k, -\pi/2+\epsilon, \pi/2-\epsilon),
\qquad
D'_k = \frac{1}{\mathrm{softplus}(\rho_{k} + \lambda_z\Delta z_k)+\epsilon},
\end{equation}
where $\rho_{k}=\mathrm{softplus}^{-1}(D_{k}^{-1})$ is the unconstrained parameter corresponding to the base inverse depth, and $\lambda_{xy},\lambda_z$ control the residual magnitude. The corrected center is then
\begin{equation}
\mu_k = D'_k\mathbf{d}(\theta'_k,\phi'_k).
\end{equation}
This angular composition makes the learned displacement consistent with the ERP sampling grid and preserves the horizontal $S^1$ closure of the panorama.

Other attributes follow the same residual composition. Scale is applied as a bounded multiplicative update to the latitude-corrected base scale, quaternions are residual updates to the identity rotation, colors are initialized from the panorama, and opacities from a constant prior; the corresponding activations keep scales positive and colors and opacities in valid ranges. During training, these residuals provide local corrections while keeping scales positive, colors in range, and opacities valid. The final output is flattened into a standard 3DGS set and can be rendered by an off-the-shelf splatting renderer without special panoramic logic.

\subsubsection{Training objective.}
We train Stage~II on HM3D dataset~\cite{ramakrishnan2021hm3d} using differentiable rendering. For each training scene, the model takes an ERP panorama and its depth as input, predicts a 3DGS scaffold, and renders it into the target view. Let $R(\mathcal{G})$ denote the rendering of the predicted Gaussians at the target camera, and let $I^{tgt}$ be the corresponding ground-truth image. The main reconstruction objective combines pixel, perceptual, and structural terms:
\begin{equation}
\mathcal{L}_{\mathrm{tgt}}
= \lambda_1\lVert R(\mathcal{G})-I^{tgt}\rVert_1
+ \lambda_{\mathrm{lpips}}\operatorname{LPIPS}(R(\mathcal{G}), I^{tgt})
+ \lambda_{\mathrm{ssim}}\bigl(1-\operatorname{SSIM}(R(\mathcal{G}), I^{tgt})\bigr).
\end{equation}
This term trains the scaffold to be directly renderable from target viewpoint. To keep the scaffold anchored to the observed input, we also render it back to the source panorama or its corresponding source views and apply a source reconstruction loss. For geometric stability, we include an inverse-depth loss when depth supervision is available:
\begin{equation}
\mathcal{L}_{\mathrm{depth}}
= \lVert D^{-1}_{\mathrm{pred}}-D^{-1}_{\mathrm{gt}}\rVert_1.
\end{equation}
We further regularize the angular--depth position residuals,
\begin{equation}
\mathcal{L}_{\Delta}
= \operatorname{mean}\bigl(\max(|\Delta_{\theta\phi z}|-\delta,0)\bigr),
\end{equation}
which allows small local corrections but discourages large deviations from the depth-initialized scaffold. The final objective is
\begin{equation}
\mathcal{L}_{\mathrm{GS}}
= \mathcal{L}_{\mathrm{tgt}}
+ \lambda_{\mathrm{src}}\mathcal{L}_{\mathrm{src}}
+ \lambda_{\mathrm{depth}}\mathcal{L}_{\mathrm{depth}}
+ \lambda_{\Delta}\mathcal{L}_{\Delta}.
\end{equation}
This objective matches the intended role of Stage~II: the scaffold should be faithful to the input panorama, stable as geometry, and directly renderable from novel viewpoints.
\subsection{Stage III: Gaussian-Conditioned Video Rendering}\label{sec:stage3}

Stage~III acts as a renderer over the Gaussian scaffold. Given a user-specified camera trajectory $\{\pi_t\}_{t=1}^{T}$, the Stage~II scaffold is first rendered by 3D Gaussian splatting into a controllable conditioning stream,
\begin{equation}
\hat{V}_t = \mathrm{Render}(\mathcal{G}, \pi_t),
\qquad
\hat{V}_{1:T}=\{\hat{V}_t\}_{t=1}^{T}.
\end{equation}
The rendered frames inherit the key advantages of the explicit scaffold: they follow the requested camera motion, preserve a persistent scene layout, and remain spatially anchored over long trajectories. However, they are still imperfect as final observations. A feed-forward panoramic Gaussian scaffold may contain floaters in depth-ambiguous regions, grazing-angle aliasing on floors and ceilings, disocclusion holes when the camera moves away from the input pose, and residual temporal splatting artifacts under sub-pixel motion. Stage~III converts this dense but imperfect RGB stream $\hat{V}_{1:T}$ into the final video $V_{1:T}=\{V_t\}_{t=1}^{T}$.

This design closes the three-stage factorization. Stage~I has already completed the missing field of view, and Stage~II has already anchored that completion into an explicit 3DGS scene. Stage~III therefore does not need to hallucinate a world from scratch. Instead, it enhances appearance and repairs local rendering artifacts while respecting the camera motion and scene layout implied by the scaffold. In this sense, MoVerse differs from purely implicit video world models: long-range spatial memory is stored in $\mathcal{G}$, while the video model performs local, high-quality observation rendering.

\begin{figure}[t]
  \centering
  \includegraphics[width=\linewidth]{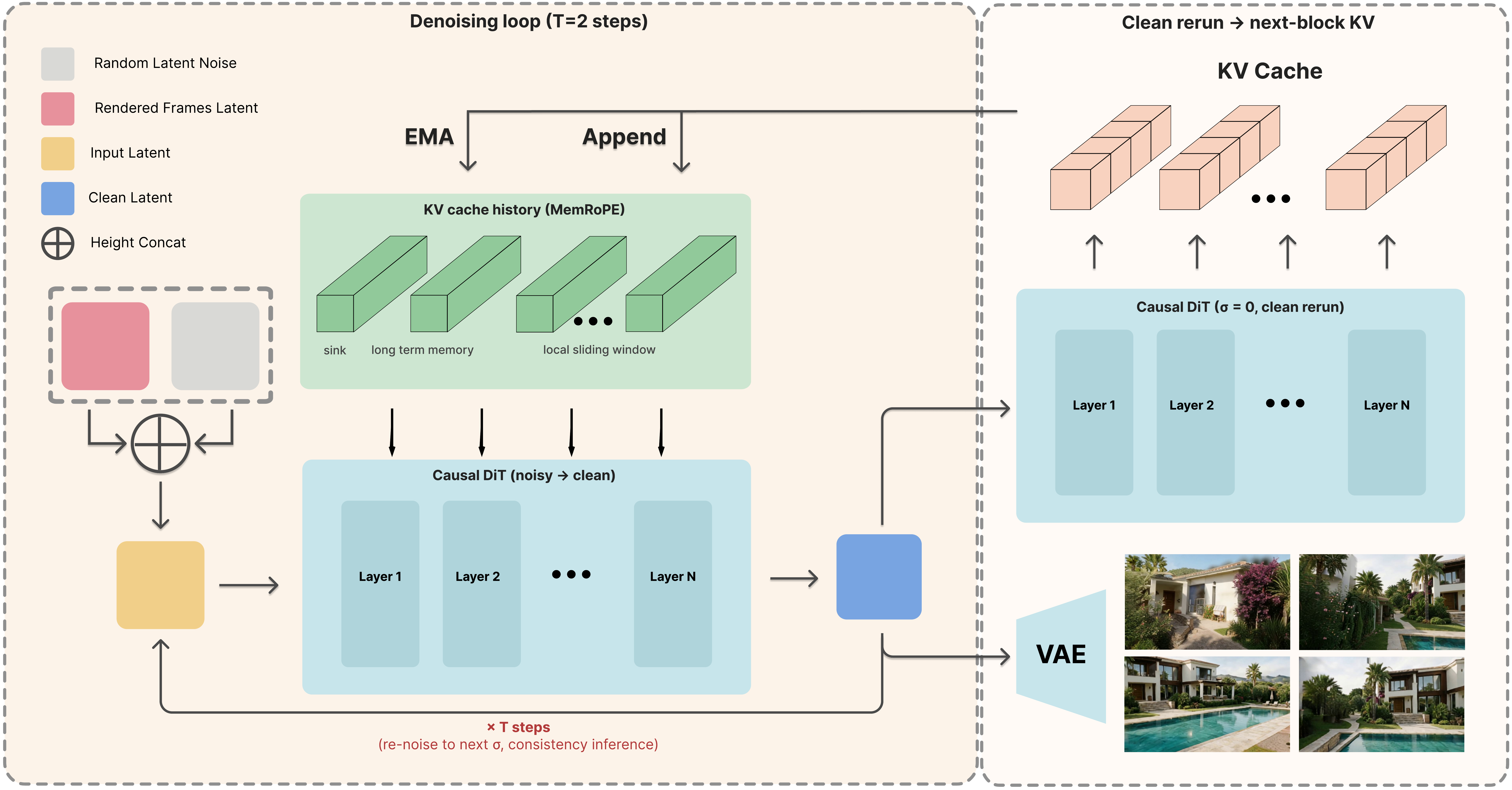}
  \caption{\textbf{Stage~III Real Time Rendering.} The rendered image latents are passed through a causal auto-rergessive DiT to generate clean images. We use MemRoPE to manage KV Cache for infinite exploration. }
  \label{fig:stage3_framework}
\end{figure}

\subsubsection{Bidirectional conditional teacher.}
We first adapt a modern text-to-video diffusion transformer, initialized from Wan2.1-T2V-1.3B~\cite{wan2025wan}, into a bidirectional Gaussian-conditioned video renderer. The teacher models
\begin{equation}
p(V_{1:T}\mid \hat{V}_{1:T}, \tau),
\end{equation}
where $\tau$ is a text prompt. The prompt provides a weak semantic prior, while the Gaussian-rendered RGB stream is the dominant spatial condition. This prevents the video generator from behaving as a free text-to-video model: its role is to translate the scaffold rendering into a cleaner observation, not to replace the scaffold.

Let $z\in\mathbb{R}^{C\times T\times H\times W}$ be the VAE latent of the target video $V_{1:T}$ and $c\in\mathbb{R}^{C\times T\times H\times W}$ be the VAE latent of the Gaussian-rendered conditioning video $\hat{V}_{1:T}$. At diffusion time $t$, the noisy target latent is
\begin{equation}
z_t = (1-\sigma_t)z + \sigma_t\epsilon,
\qquad \epsilon\sim\mathcal{N}(0,I).
\end{equation}
We concatenate the noisy target tokens and conditioning tokens as
\begin{equation}
\tilde{z} = [\,z_t \;\Vert\; c\,].
\end{equation}
The two halves use shared rotary positional indices: every conditioning token receives the same spatial--temporal position code as its aligned target token. This shared-RoPE design makes the condition and target collide at the same location in the transformer coordinate system, encouraging the model to treat $c$ as a dense aligned rendering condition rather than as a separate video placed at an offset position.

The teacher is trained with a flow-matching objective, evaluated only on the target half of the concatenated tokens:
\begin{equation}
\mathcal{L}_{\mathrm{teacher}}
= \mathbb{E}_{z,c,\tau,\sigma_t,\epsilon}
\left[
\left\|v_{\theta}(\tilde{z},\tau,t) - (\epsilon-z)\right\|_2^2
\right],
\end{equation}
where $v_{\theta}$ denotes the diffusion transformer prediction. The bidirectional teacher can attend to the full clip, so it serves as a quality model for Gaussian-conditioned video rendering rather than as the final deployment model.

\subsubsection{Causal student for streaming interaction.}
The bidirectional teacher is unsuitable for interactive roaming because it requires access to the full video window and performs multi-step generation over all frames jointly. We therefore distill it into a causal autoregressive student. The student generates one latent block at a time, conditions on the current Gaussian-rendered frame block, and attends to recently generated history through a local key--value cache. This converts the teacher into a bounded-latency renderer that can stream frames as the user changes the camera trajectory.

The distillation follows the Self Forcing~\cite{huang2026self} and DMD~\cite{yin2023one} principle.
During training, the student is rolled out autoregressively and is exposed to its own generated context, so it learns to remain stable under the same causal conditions used at inference. A short bidirectional-to-causal warm-up~\cite{hong2025relic} initializes the student before distribution matching. Within distribution matching, we adopt RAVEN~\cite{lu2026raven}: each self-rollout is repacked into a teacher-forcing sequence of clean endpoints and noisy intermediates, so the DMD gradient also flows through the clean half's QKV projections and supervises how the student encodes its own past context into the KV cache used by future blocks. The resulting model trades global bidirectional refinement for low-latency local refinement. This trade-off is appropriate for MoVerse because the explicit scaffold, not the video model cache, carries long-range scene consistency.

\subsubsection{Training data mixture.}
Training pairs are organized as triplets $(V^{\mathrm{gt}}_{1:T}, \hat{V}_{1:T}, \tau)$, where $V^{\mathrm{gt}}_{1:T}$ is a target video, $\hat{V}_{1:T}$ is a paired geometry-conditioned rendering, and $\tau$ is a caption. The purpose of the mixture is not to simply aggregate datasets, but to expose the renderer to progressively more realistic versions of the signal it will receive at deployment: camera-aligned but incomplete geometric projections, genuine Gaussian splatting artifacts, and finally renderings from panorama-conditioned scaffolds.

The first family uses real videos with known or recoverable camera motion to synthesize trajectory-aligned proxy renderings from a single reference view. We estimate depth for the reference frame, lift it once into a world-space point set, and reproject it through the remaining cameras. The resulting conditioning videos preserve the target trajectory and coarse scene layout, but they naturally contain missing regions, broken boundaries, and many-to-one warp artifacts. These examples are useful precisely because they isolate the basic inpainting problem caused by camera translation: the renderer must complete newly exposed regions while respecting the motion and layout encoded by the geometric projection.

The second family replaces point reprojection with rendered Gaussian scaffolds constructed from sparse real-video keyframes. For each scene, several keyframe-level Gaussian predictions are brought into a common metric frame, merged, and rendered along the original camera path. These conditioning videos are no longer generic warps; they are actual splatting renderings, and therefore expose the model to the failure modes of Gaussian representations: depth-dependent floaters, imperfect opacity accumulation, anisotropic support errors, and aliasing on grazing surfaces. This family teaches the video model to act as a renderer for explicit Gaussian geometry rather than merely as a video inpainting model.

The third family matches the MoVerse interface most closely. Instead of starting from a NFOV image, we start from panoramic observations, construct a Gaussian scaffold with the same panorama-to-Gaussian pathway used by Stage~II, and render that scaffold along immersive camera trajectories. We build such pairs from UE-rendered panoramas and additional panoramic sources, including LayerPano3D~\cite{yang2025layerpano3d}, Matterport3D~\cite{chang2017matterport3d}, and Polyhaven HDRIs~\cite{polyhaven}. These samples provide the strongest supervision for the deployment-time interface because the condition $\hat{V}_{1:T}$ is produced by rendering a panorama-conditioned scaffold. They are complemented by the first two families, which provide broader scene diversity and denser coverage of generic disocclusion and Gaussian-rendering artifacts.

The mixture therefore follows a representation-driven progression:
\begin{equation}
\text{trajectory-aligned geometric proxy}
\;\rightarrow\;
\text{explicit Gaussian render}
\;\rightarrow\;
\text{panorama-conditioned Gaussian render}.
\end{equation}
This curriculum first teaches robust artifact repair under camera motion, then specializes the renderer to Gaussian splatting artifacts, and finally adapts it to the Stage~II-to-Stage~III interface used by MoVerse at deployment.

\subsubsection{Real-time inference.}
At deployment, the distilled student runs causally with $K=1$ latent frame per autoregressive block. Under the Wan VAE temporal stride, each latent block corresponds to four pixel-space frames. We use a two-step denoising schedule and a MemRoPE-style local key--value cache~\cite{kim2026memrope}: one sink frame, one long-term EMA memory token, and a sliding window of three local frames, with online RoPE indexing that keeps temporal positions inside the trained range over long rollouts. This cache maintains short-horizon temporal continuity, while the Stage~II scaffold anchors long-horizon spatial consistency. The standard VAE decoder is replaced by a TAEHV decoder~\cite{BoerBohan2025TAEHV} for faster streaming.

These choices deliberately trade global bidirectional refinement for bounded-latency local refinement. Since every output block is conditioned on the current scaffold rendering, the model does not need to remember the entire world in its hidden state. In our deployment configuration, this causal renderer reaches 8~FPS end-to-end roaming of the scene on a single NVIDIA RTX~4090 GPU.
\section{Results}
\label{sec:results}

In this section, we present qualitative results for MoVerse across the full pipeline and its three stages.

\begin{figure}[!htbp]
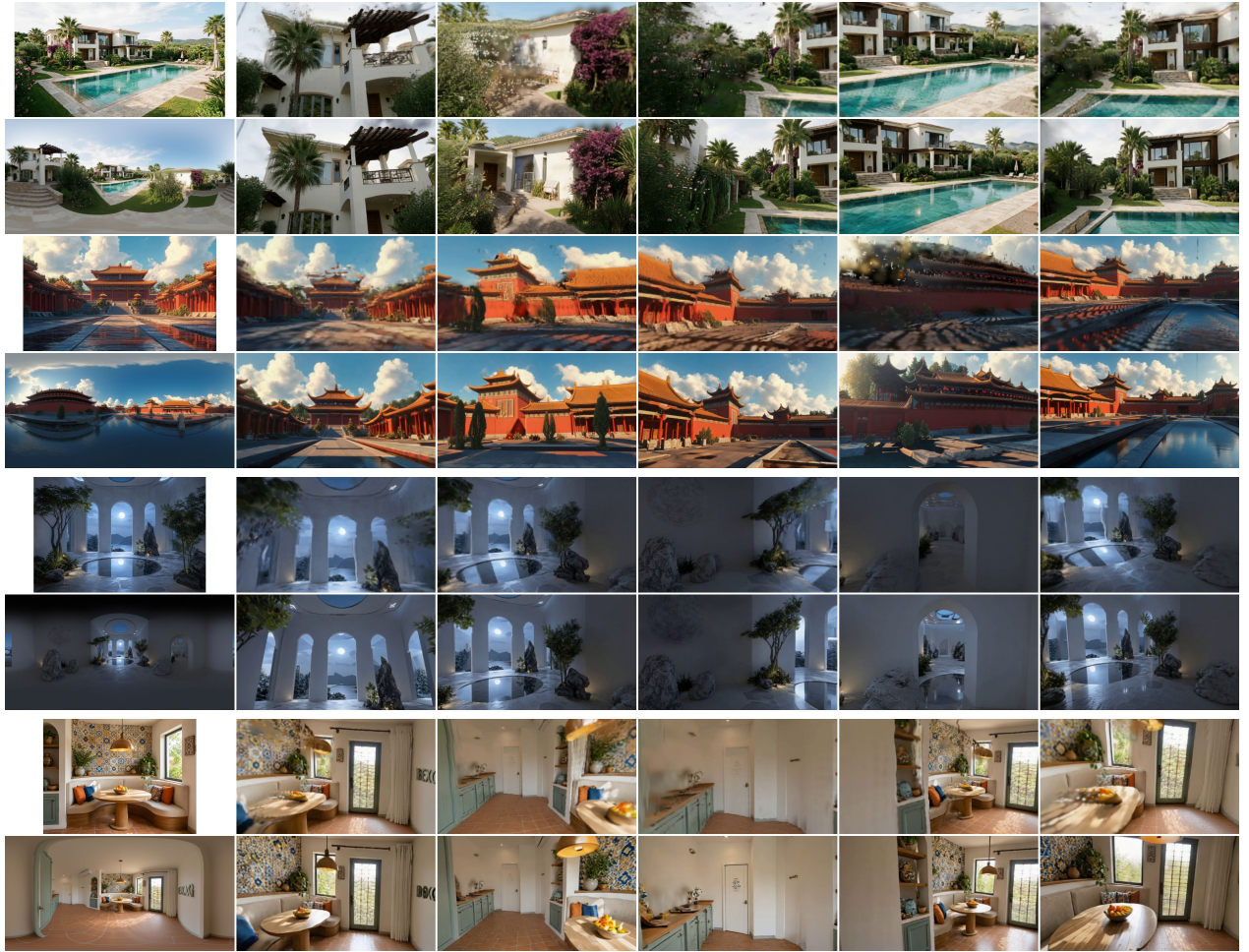

  \centering
  \includegraphics[width=\linewidth]{results/full_pipeline/scene1.pdf}
  \vspace{1mm}
  \includegraphics[width=\linewidth]{results/full_pipeline/scene2.pdf}
  \vspace{1mm}
  \includegraphics[width=\linewidth]{results/full_pipeline/scene3.pdf}
  \vspace{1mm}
  \includegraphics[width=\linewidth]{results/full_pipeline/scene4.pdf}
  \caption{\textbf{Full pipeline results.} For each scene, the first column shows the input NFOV image and the Stage~I gravity-aligned ERP panorama. The remaining columns show Stage~II Gaussian scaffold renderings and the corresponding Stage~III autoregressive video-rendering outputs along the same camera trajectory.}
  \label{fig:results_full_pipeline}
\end{figure}

\begin{figure}[!htbp]
  \centering
  \includegraphics[width=0.9\linewidth]{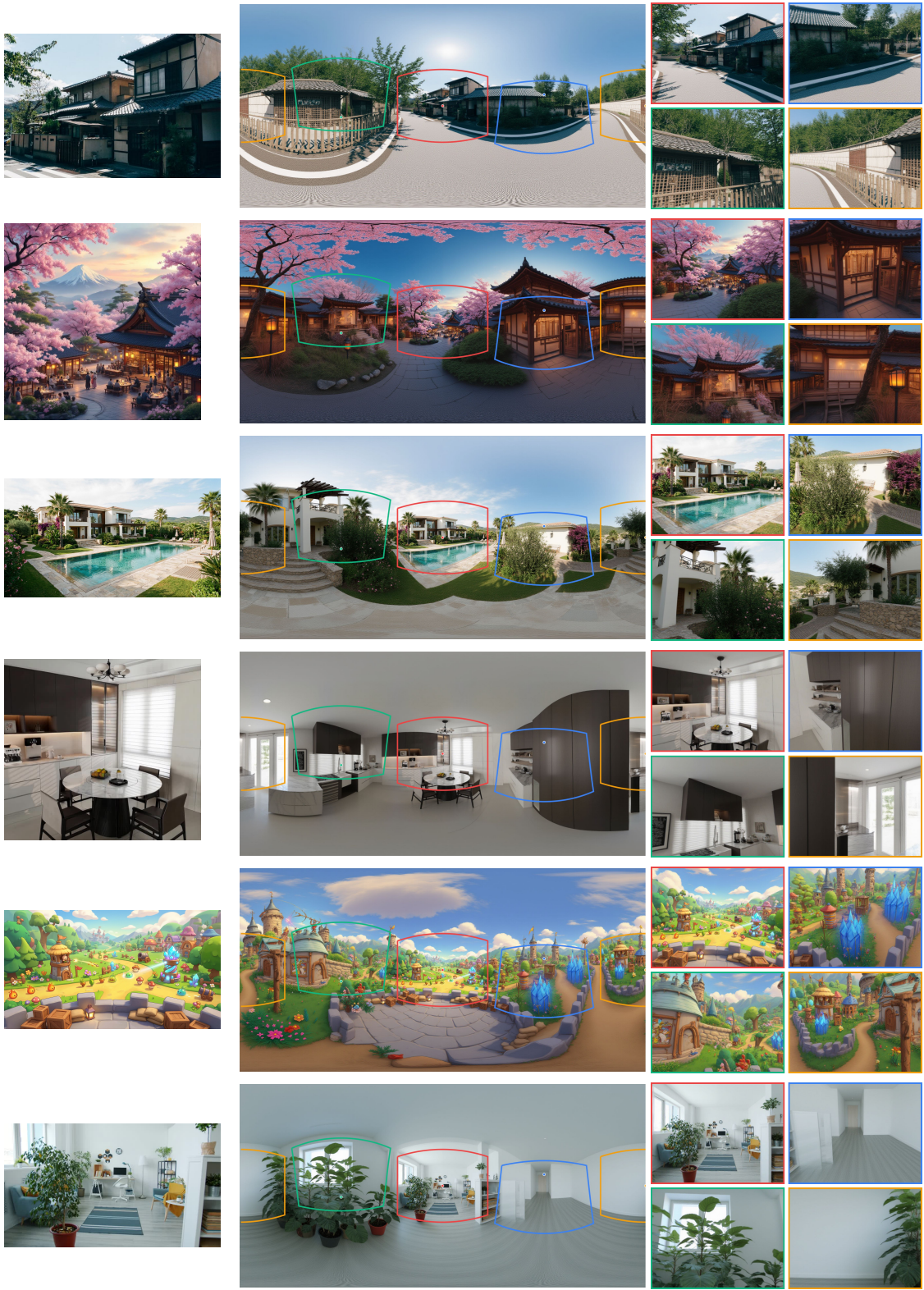}
  \caption{\textbf{Stage~I panoramic generation results.} Each example shows a perspective NFOV input, the completed gravity-aligned ERP panorama, and perspective views rendered from the panorama, including seam-crossing views.}
  \label{fig:results_stage1}
\end{figure}

\begin{figure}[!htbp]
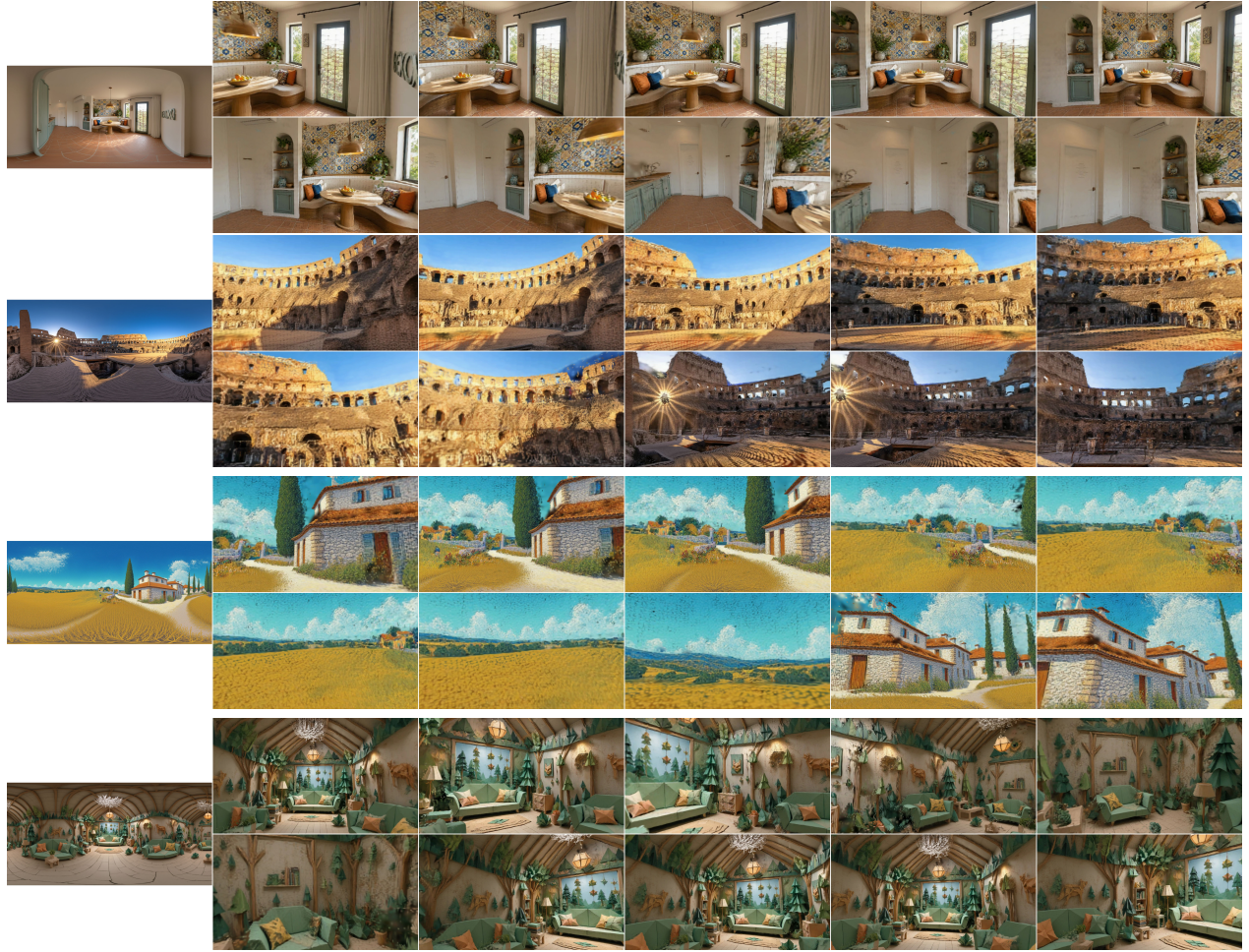

  \centering
  \includegraphics[width=\linewidth]{results/stage2/alcove.pdf}
  \vspace{1mm}
  \includegraphics[width=\linewidth]{results/stage2/colosseum.pdf}
  \vspace{1mm}
  \includegraphics[width=\linewidth]{results/stage2/farm.pdf}
  \vspace{1mm}
  \includegraphics[width=\linewidth]{results/stage2/paper.pdf}
  \caption{\textbf{Stage~II Gaussian scaffold rendering results.} For each scene, the first column shows the input condition, and the remaining columns show novel-view frames rendered directly from the 3D Gaussian scaffold along two camera trajectories.}
  \label{fig:results_stage2}
\end{figure}

\begin{figure}[!htbp]
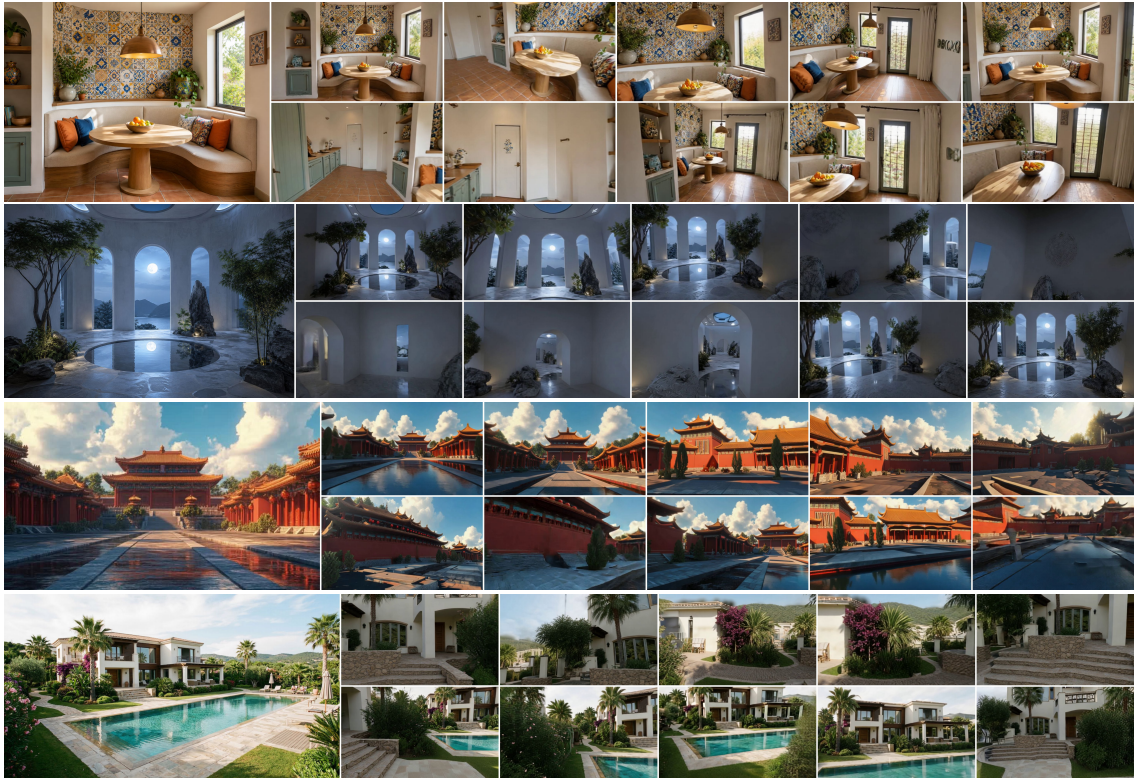

  \centering
  \includegraphics[width=0.9\linewidth]{results/stage3/alcove_input_gen.pdf}
  \vspace{0.3mm}
  \includegraphics[width=0.9\linewidth]{results/stage3/moon_input_gen.pdf}
  \vspace{0.3mm}
  \includegraphics[width=0.9\linewidth]{results/stage3/palace_input_gen.pdf}
  \vspace{0.3mm}
  \includegraphics[width=0.9\linewidth]{results/stage3/villa2_input_gen.pdf}
  \caption{\textbf{Stage~III autoregressive rendering results.} For each scene, the first column shows the Gaussian-rendered input condition, and the remaining columns show frames generated by the causal autoregressive renderer along two camera trajectories.}
  \label{fig:results_stage3}
\end{figure}

\begin{figure}[!htbp]
  \centering
  \includegraphics[width=0.9\linewidth]{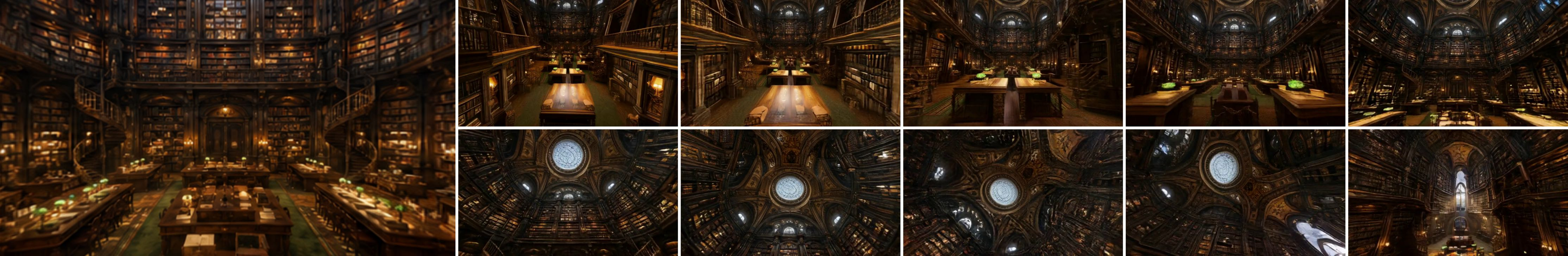}
  \vspace{0.3mm}
  \includegraphics[width=0.9\linewidth]{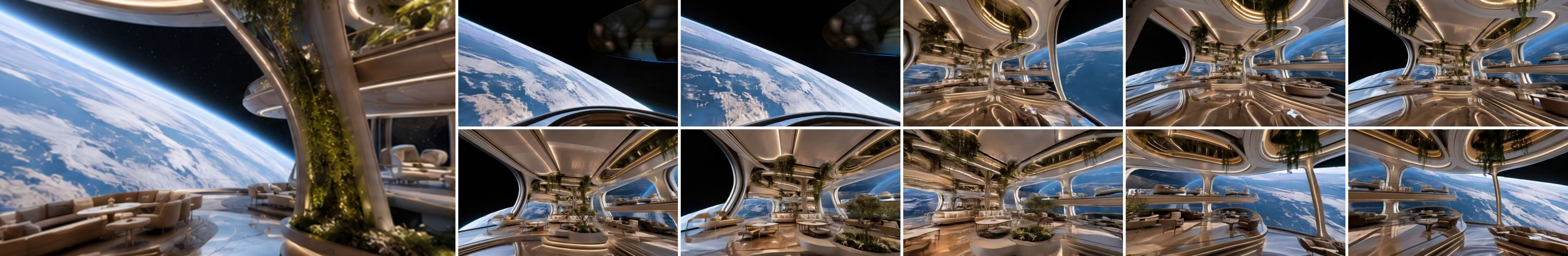}
  \vspace{0.3mm}
  \includegraphics[width=0.9\linewidth]{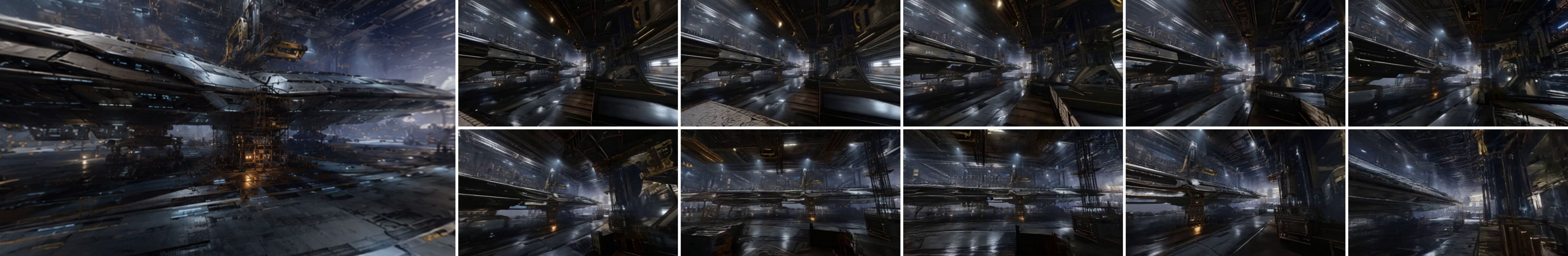}
  \vspace{0.3mm}
  \includegraphics[width=0.9\linewidth]{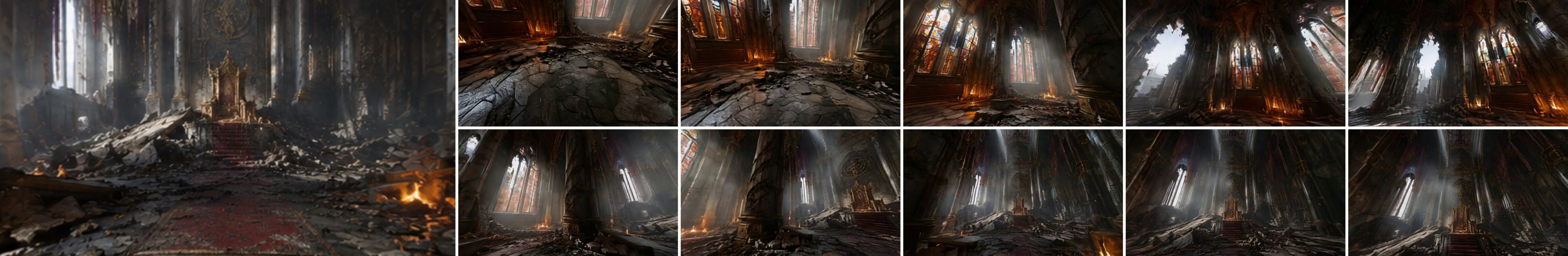}
  \caption{\textbf{Bidirectional conditional teacher results.} For each scene, the first column shows the Gaussian-rendered input condition, and the remaining columns show frames synthesized by the bidirectional Gaussian-conditioned video renderer along two camera trajectories.}
  \label{fig:results_bidirectional}
\end{figure}

\subsection{Full pipeline results}
Fig.~\ref{fig:results_full_pipeline} shows end-to-end results produced from a single NFOV image. Each example contains the input image, the gravity-aligned ERP panorama from Stage~I, Gaussian-rendered conditioning frames from the Stage~II 3D Gaussian scaffold, and the final observations generated by the Stage~III causal autoregressive renderer under the same camera trajectory. The results illustrate the complementary roles of the explicit scaffold and the learned renderer: the 3D Gaussian scaffold provides camera-controllable spatial structure, while the Gaussian-conditioned video renderer improves perceptual quality and temporal continuity.

\subsection{Stage I: Panoramic Generation}
Fig.~\ref{fig:results_stage1} evaluates the Stage~I panoramic generation module. Given a perspective NFOV input, Stage~I synthesizes a gravity-aligned ERP panorama that can be reprojected into multiple perspective views. We include seam-crossing views to examine whether the generated panorama remains coherent under the horizontal $S^1$ topology of the ERP domain.

\subsection{Stage II: Gaussian Scene Generation}
Fig.~\ref{fig:results_stage2} visualizes trajectories rendered directly from the Stage~II 3D Gaussian scaffold. The scaffold produces camera-controllable Gaussian-rendered observations that preserve the global layout of the completed panorama.

\subsection{Stage III: Gaussian-Conditioned Video Rendering}
Fig.~\ref{fig:results_stage3} shows outputs from the autoregressive video renderer. The renderer translates Gaussian-rendered conditioning frames into final video observations while streaming along user-specified camera trajectories. Across the shown trajectories, it enhances the scaffold renderings into visually coherent observations while preserving the scene layout imposed by the explicit representation.

\subsection{Bidirectional Conditional Teacher Results}
Fig.~\ref{fig:results_bidirectional} shows results from the bidirectional conditional teacher. The teacher is implemented as a bidirectional Gaussian-conditioned video renderer and can attend to the full temporal window, serving as a quality model for Gaussian-conditioned video rendering. These results provide the visual target used to guide distillation of the causal autoregressive student.

\section{Discussion}
MoVerse explores a hybrid route between explicit 3D reconstruction and implicit video world modeling. Its central design choice is to store long-range spatial memory in a persistent panoramic 3D Gaussian scaffold, while using a causal video renderer only for local, high-fidelity observation synthesis. This separation is useful in practice: the scaffold fixes the camera trajectory and scene layout over long horizons, and the video model can focus on repairing splatting artifacts, filling small disocclusions, and improving perceptual realism. The three-stage factorization also makes the system modular. Improvements in panorama generation, feed-forward Gaussian prediction, or video distillation can be incorporated independently without changing the overall interface between stages.

At the same time, this factorization exposes several limitations. First, the final world is bounded by the quality of Stage~I panoramic completion. If the generated panorama introduces semantically inconsistent rooms, incorrect horizon structure, or implausible content behind the camera, Stage~II will faithfully lift those errors into the scaffold, and Stage~III may render them more convincingly rather than correct them. Second, the Gaussian scaffold remains an approximate geometric representation. Depth ambiguity, thin structures, reflective surfaces, transparent objects, and regions near the poles of the ERP domain can still produce floaters, holes, or unstable opacity accumulation under large translations. Third, the causal video renderer trades global clip-level optimization for streaming latency. Although the explicit scaffold provides long-range spatial consistency, the renderer may still introduce short-term texture drift, over-smoothing, or delayed correction when the conditioning render is severely degraded. Finally, the current system depends on a staged training pipeline and multiple data sources, including canonical panoramas, depth-supervised Gaussian prediction, and paired scaffold-render/video data; simplifying this supervision remains an important direction for broader deployment.

Future work should strengthen the feedback between stages rather than treating them as a one-way pipeline. For example, uncertainty from the panoramic generator could guide Gaussian density allocation, scaffold rendering errors could trigger local panorama or geometry refinement, and the video renderer could expose consistency signals back to the scaffold. Another promising direction is to extend the static-scene assumption toward dynamic objects and editable worlds, where the persistent scaffold must support object-level manipulation as well as camera motion. Overall, MoVerse shows that real-time video world modeling need not choose between explicit geometry and generative video. A dense panoramic Gaussian scaffold can provide durable spatial memory and controllability, while a distilled autoregressive renderer supplies interactive visual quality, enabling single-image world creation with real-time roaming.

\section*{Acknowledgements}
We thank Jie Ma and Ben Hu for their valuable assistance with data collection, curation, and preprocessing.

{
\small
\bibliographystyle{IEEEtran}
\bibliography{main}
}


\end{document}